\documentclass[10pt,twocolumn,letterpaper]{article}

\usepackage{cvpr}

\usepackage{url}
\usepackage{times}
\usepackage{epsfig}
\usepackage{graphicx}
\usepackage{amsmath}
\usepackage{xfrac}
\usepackage{amssymb}
\usepackage{graphicx}
\usepackage{amsmath}
\usepackage{xfrac}
\usepackage{amssymb}
\usepackage{bm}
\usepackage{booktabs}
\usepackage{caption}
\usepackage{mathrsfs}
\usepackage{tabulary,multirow,overpic,xcolor}
\usepackage{cuted}
\usepackage[caption=false]{subfig}
\usepackage[american]{babel}

\newcolumntype{x}[1]{>{\centering\arraybackslash}p{#1pt}}

\newcommand{\app}{\raise.17ex\hbox{$\scriptstyle\sim$}}

\newlength\savewidth

\usepackage[pagebackref=true,breaklinks=true,letterpaper=true,colorlinks,bookmarks=false]{hyperref}

\cvprfinalcopy 

\def\cvprPaperID{1760} 

\ifcvprfinal\fi
\begin{document}

\title{Learnable Higher-order Representation for Action Recognition}

\author{Kai Hu\\
Carnegie Mellon university\\
{\tt\small kaihu@cs.cmu.edu}
\and
Bhiksha Raj\\
Carnegie Mellon university\\
{\tt\small bhiksha@cs.cmu.edu}
}

\maketitle

\begin{abstract}
Capturing spatiotemporal dynamics is an essential topic in video recognition. In this paper, we present learnable higher-order operations as a generic family of building blocks for capturing spatiotemporal dynamics from RGB input video space. Similar to higher-order functions, the weights of higher-order operations are themselves derived from the data with learnable parameters. Classical architectures such as residual learning and network-in-network are first-order operations where weights are directly learned from the data. Higher-order operations make it easier to capture context-sensitive patterns, such as motion. Self-attention models are also higher-order operations, but the attention weights are mostly computed from an affine operation or dot product. The learnable higher-order operations can be more generic and flexible. Experimentally, we show that on the task of video recognition, our higher-order models can achieve results on par with or better than the existing state-of-the-art methods on  Something-Something (V1 and V2), Kinetics and Charades datasets.
\end{abstract}
\section{Introduction}
\begin{figure*}[htbp]
\centering
\subfloat[pull something from left to right\label{fig_one1}]{
\begin{minipage}[t]{0.245\linewidth}
\centering
\includegraphics[width=.95\textwidth]{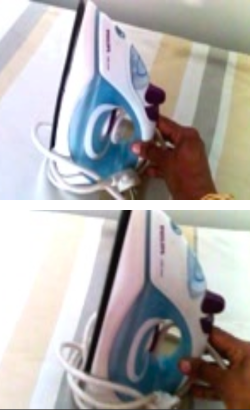}
\end{minipage}%
}%
\subfloat[push something from right to left\label{fig_one2}]{
\begin{minipage}[t]{0.245\linewidth}
\centering
\includegraphics[width=.95\textwidth]{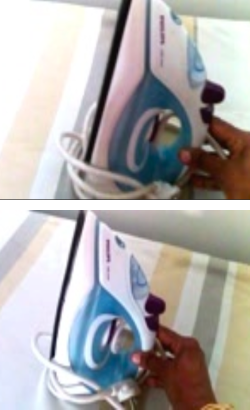}
\end{minipage}%
}
\subfloat[pull something from right to left]{
\begin{minipage}[t]{0.245\linewidth}
\centering
\includegraphics[width=.95\textwidth]{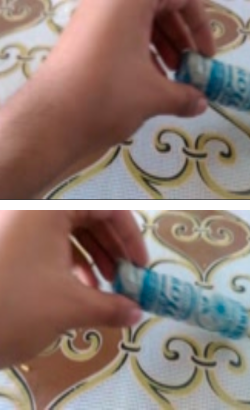}
\end{minipage}
}%
\subfloat[push something from left to right\label{fig_one4}]{
\begin{minipage}[t]{0.245\linewidth}
\centering
\includegraphics[width=.95\textwidth]{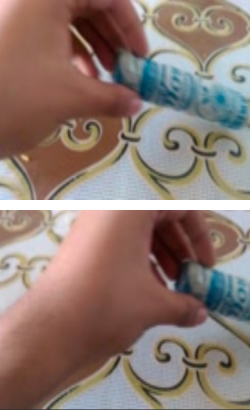}
\end{minipage}
}%
\centering
\caption{Different contexts of the hand in four different categories. In Figure \ref{fig_one1}, since the hand moves from left to right and the hand is on the right side of the iron, it is $\emph{pull from left to right}$. Figure \ref{fig_one4} has the same hand movement, but it is a different category since the hand is on the left of the pen. Figure \ref{fig_one2} is a reverse action of Figure \ref{fig_one1}, but it is not $\emph{pull from right to left}$. Let A be a boolean that the hand moves from left to right and B be a boolean that the hand is on the right of the object. The action $\emph{push}$ is A XOR B.}
\label{fig_one}
\end{figure*}
Actions in videos arise from motions of objects with respect to other objects and/or the background. To understand an action, an effective architecture should recognize not only the appearance of the target object associated with the action, but also how it relates to other objects in the scene, in both space and time. Figure \ref{fig_one} shows four different categories of actions. Each column shown an action where, in temporal order, the figures above occur before the figures below. Recognizing the hand and the object is not enough. To distinguish  $\emph{left to right}$ motion from $\emph{right to left}$ motion, the model must know how the hand moves against the background. It is more complicated to classify $\emph{pull}$ and $\emph{push}$ since it is an XOR operation on the relative positions of the hand and the object resulting from the hand's movements.

The key point here is the need for recognizing patterns in spatiotemporal context. Even the same hand-iron-background combination has different meanings in different spatiotemporal contexts. The number of combinations increases sharply as scenes become more complicated and the number of objects involved increases. Conventional convolution recognizes fixed patterns determined by the fixed filter parameters, so it is difficult to capture various variations that distinguish the action categories. To recognize {\em every} object-in-context pattern, the model needs to have more detailed filters, potentially leading to a blow up of the number parameters. 

On the other hand, although the object-in-context patterns can vary, they are related through a higher-order structure: pushing an iron,pulling an iron, pushing a pen, and so on. 
We hypothesize that the structure of object-in-context patterns can be learned, i.e., the model can learn to conclude object-in-context pattern given the context, and propose a corresponding feature extractor. 

Explicitly, let $\bm{X}$ and $\bm{Y}$ respectively represent the input and output of a convolution. Let $\bm{y}_p$ and $\{\bm{x}_{p'}\}$ represent a specific position of $\bm{Y}$ and the set of positions of $\bm{X}$ where $\bm{y}_p$ is computed, respectively. Denote conventional convolution operation (first-order) as $\bm{Y} = f(\bm{X}; \Theta)$ where $\Theta$ is the shared parameters at different positions. The parameters act as determined feature extractors as $\bm{y}_p = f(\{\bm{x}_{p'}\}; \Theta)$ for different positions.

As we analyze, the visual pattern of the target object can vary in different contexts, and feature extractors (filters) that ignore this dependence are not optimal. We replace the fixed filters with context-dependent filters $\bm{y}_p = f(\{\bm{x}_{p'}\}; \bm{w}_p)$ where the filters $\bm{w}_p$ are in turn obtained as $\bm{w}_p = g(\{\bm{x}_{p''}\}; \Theta)$. The mapping $g$ is the structure of object-in-context patterns and $\Theta$ are the learned parameters as we hypothesize. The entire relation between $\bm{Y}$ and $\bm{X}$ can be  represented through $\bm{Y} = f(\bm{X}; g(\bm{X}; \Theta))$. We define this as a higher-order operation since the function $f$ takes function $g$ as as an argument.

The proposed model is able to capture spatiotemporal contexts effectively. We test our method on four benchmark datasets for action recognition: Kinetics-400 \cite{i3d}, Something-Something V1 \cite{sth-sth-v2}, Something-Something V2, and Charades datasets \cite{charades}. Specifically, we make comprehensive ablation studies on Something-Something V1 datasets  and further evaluate on the other three datasets to demonstrate the generality of our proposed method. The experiments establish significant advantages of the proposed models over existing algorithms, achieving results on par with or better than the current state-of-the-art methods. 


\section{Related Work}
\textbf{Action Recognition.} 
Many action recognition methods are based on high-dimensional encodings of local features. For instance, Laptev et al \cite{laptev2008learning} used as local features histograms of oriented gradients \cite{dalal2005histograms} and histograms of optical flow as sparse interest points. The features are encoded into a bag of features representation. 
Wang et al \cite{wang2011action} and Peng et al \cite{peng2014action} made use of dense point trajectories which are computed using optical flow.
The high performance of 2D ConvNets in image classification tasks \cite{imagenet} makes it appealing to try to reuse them for video recognition tasks. Tran et al investigated 3D ConvNets to learn spatiotemporal features end-to-end \cite{c3d}. Some researchers tried to save computation by replacing 3D convolutions with separable convolutions \cite{p3d, r2plus1d} or mixed convolutions \cite{r2plus1d, s3d}. Meanwhile, Carreira and Zisserman introduced an inflation operation \cite{i3d}. It allows for converting pre-trained 2D models into 3D. Simonyan et al designed a two-stream architecture to capture appearance and motion information separately \cite{twostream}. The spatial stream uses RGB frames as inputs, while the temporal stream learns from stacked optical flow. Wang et al further generalized this framework to learn long-range dependencies by temporal segment \cite{tsn}.  Self-attention mechanisms have recently been successfully applied in visual recognition \cite{sharma2015action,girdhar2017attentional,miech2017learnable,Wang_nonlocalCVPR2018,baradel2018human,Girdhar_2019_CVPR}. Though we do not use key and query pairs, our method can be seen as a more generalized form of self-attention that learns more structured information from the feature map.

\textbf{Spatiotemporal Context.}
Contextual information is very important for action recognition. \cite{galleguillos2010context} review different approaches
of using contextual information in the field of object recognition. Several methods \cite{marszalek2009actions,sun2009hierarchical,kovashka2010learning,vail2007conditional,cao2015spatio,chen2014actionness} exploit contextual information to facilitate action recognition. Marszałek et al exploited the context of natural dynamic scenes
for human action recognition in video [\cite{marszalek2009actions}]. Sun et al modeled the spatio-temporal relationship between trajectories in a hierarchy of multiple levels \cite{sun2009hierarchical}. Kovashka et al proposed to learn the shapes of space-time feature neighborhoods that are most discriminative for a given action category \cite{kovashka2010learning}. Conditional Random Field models have also been exploited for object and action recognition \cite{vail2007conditional,cao2015spatio,chen2014actionness,NIPS2004_2652,Wang_2018_ECCV}. Wang et al investigated the non-local mean operation to captures long-range dependency by iterative utilization of local and non-local operations\cite{Wang_nonlocalCVPR2018}. Cao et al found that the global contexts modeled by non-local network are almost the same for different query positions within an image and proposed the global context (GC) block \cite{cao2019gcnet}. Qiu et al proposed a two-path method to combine local and global representations \cite{Qiu_2019_CVPR}. Wu et al used dynamic convolutions where the kernel is a linear output of the context window to learn from the context \cite{wu2019pay}. In our experiment, we also find a relatively non-local context is important but a more global context shows diminishing return. So we choose a context field that is not global but much larger than the typical convolutional kernel size. Some related work studied using the network to generate a network \cite{deutsch2018generating,ha2016hypernetworks}. The generated network is fixed after training. Ours is estimating different weights for every input.

\section{Our Approach}
In this section, we define our higher-order model for video analysis. Our model comprises the analysis of video feature maps by a position-dependent bank of spatio-temporal filters. The filter parameters are themselves derived through a smaller network at different positions.

The description below represents {\em one layer} or {\em block} of a larger model. We will refer to such second (or more generally, higher) order blocks as {\em H-blocks}. We note that the larger model may be composed entirely of H-blocks, or include H-blocks intermittently between conventional convolutional layers. To allow for this more generic interpretation we will define our blocks as working on {\em video feature maps} and producing video feature maps, where the input map may either be the original video itself or the output of prior blocks.

\subsection{Notation}
We denote the {\em input} video feature map of the H-block as $\bm{X}\in \mathbb{R}^{C_{in}\times T\times H \times W}$, where $C_{in}$ is the number of channels in each frame of the video, $T$ is the number of frames, and the height and the width of each frame are $H$ and $W$. The feature/content at position $p=(t,h,w)$, $1\leq t\leq T,1\leq h\leq W,1\leq w\leq W$, is represented as $\bm{x}_{p}$, and $\bm{x}_p \in \mathbb{R}^{C_{in}}$.

We denote the {\em output} map for the H-block as $\bm{Y}\in \mathbb{R}^{C_{out}\times T'\times H' \times W'}$.  The description below assumes, for convenience, that the spatio-temporal dimensions of the output map are identical to those of the input (i.e. $T' = T, H' = H$, and $W' = W$) although this is not essential.  Similarly to the input, we denote elements at individual spatio-temporal positions of the output as $\bm{y}_p$, where $\bm{y}_p \in \mathbb{R}^{C_{out}}$.

In our model $\bm{Y}$ is derived from $\bm{X}$ through a second-order relation of the form $\bm{Y} = f(\bm{X}, g(\bm{X}; \Theta))$ -- the relation being second order since the function $f$ relating the input and output maps takes a function $g$ as arguments to generate parameters of function $f$. Both $f(\cdot)$ and $g(\cdot)$ are convolution-like (or actual convolution) operations; hence we can use terminology drawn from convolutional neural networks to describe them. As reference, we first describe the common convolutional network structure, and subsequently build our model from it.

Following \cite{dai2017deformable}, we use a grid $\mathcal{R}$ over the input feature map to specify the receptive field size and dilation for convolution  kernels. For example (all integers below),
\begin{equation}\mathcal{R}=\left\{(t,h,w)\Big ||t|\leq K_t, |h|\leq K_h, |w|\leq K_w\right\}\end{equation}
defines a 3D kernel with kernel size $(2K_t+1)\times(2K_h+1)\times(2K_w+1)$ and dilation 1. The usual convolution operation can now be written as
\begin{equation}
\bm{y}_p = \sum_{q\in\mathcal{R}} \bm{W}_{q} \bm{x}_{p+q}.\label{eq:regularconv}
\end{equation}
where $\{\bm{W}_q,q\in\mathcal{R}\}$ are the weights of convolutional {\em filters} that scan the input $\bm{X}$. Each $\bm{W}_{q}$ is a matrix: $\bm{W}_{q}\in\mathbb{R}^{C_{out}\times C_{in}}$.  The convolution outputs are generally further processed by an activation function such as ReLU and \emph{tanh}.

Our H-block retains the same structure as above, except that the convolution operation of Equation \ref{eq:regularconv} changes to
\begin{equation}
\bm{y}_p = \sum_{q\in\mathcal{R}} \bm{W}_{p,q} \bm{x}_{p+q}.\label{eq:modifiedconv}
\end{equation}
Note that the filter $\bm{W}_{p,q}$ are now position dependent. If the model learns these filter as parameters from the data, Equation \ref{eq:modifiedconv} is a local connected layer which requires numerous parameters. In the higher-order model, filters $\bm{W}_{p,q}$ are themselves derived using an upper-level function with parameters. Representing the entire set of filter parameters as $\mathcal{W} = \{\bm{W}_{p,q}\}$, we have
\[
\mathcal{W} = g(\bm{X}; \Theta)
\]
The {\em actual} number of parameters required to define the block is the number of components in $\Theta$. We propose two models for  $g(\cdot)$ below, with different requirements for the number of parameters.

\subsection{Convolution-based second-order operation}
\begin{figure*}[htbp]
\centering
\includegraphics*[width=1\textwidth]{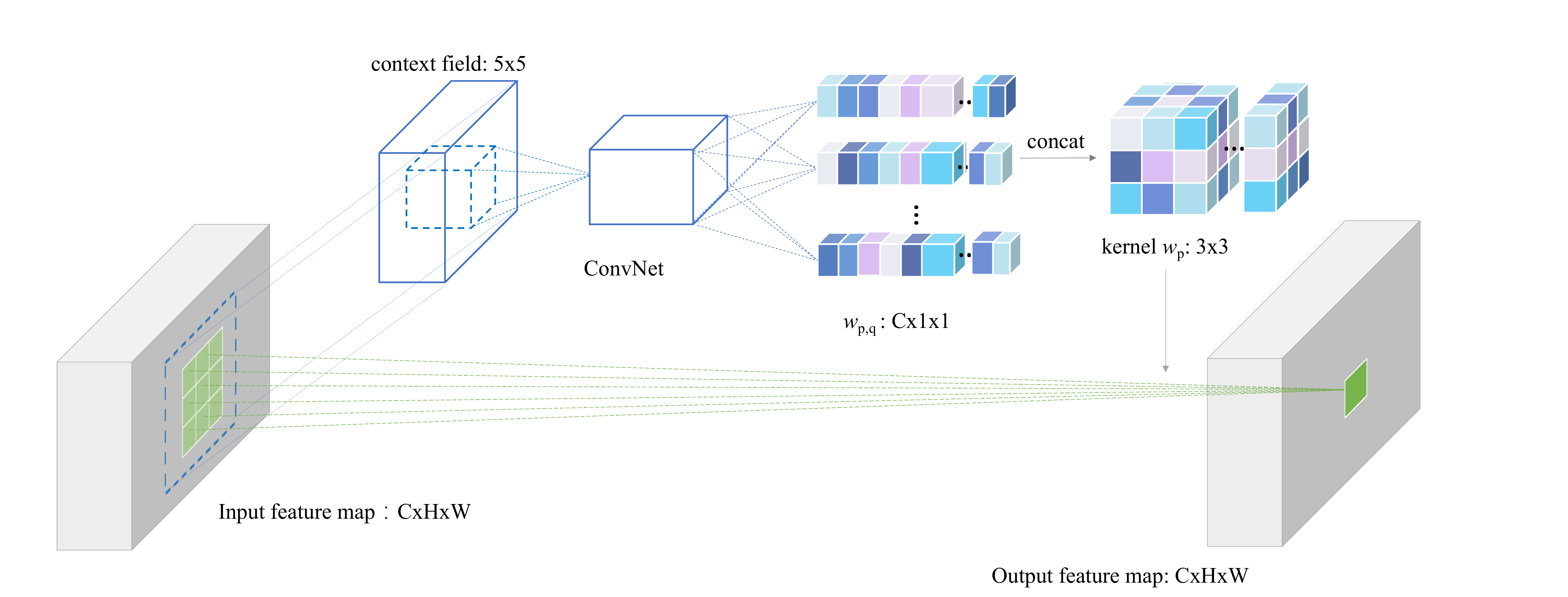}
\caption{One example of a second-operation on 2D data with channel number $C$, width $W$, and height $H$. For every position $p$ in the feature map ($HW$ positions in total), the ConvNet derives 9 C-dimensional vectors. They are concatenated into a $C\times3\times3$ filter to extract the $p^{\rm th}$ output feature from a $3 \times 3$ region of the input feature map centered at $p$.}
\label{fig:network}
\end{figure*}
In the convolution-based model for $g(\cdot)$, we derive the filter parameters $\{\bm{W}_{p,q}\}$ through a convolutional operation.  Since the total number of parameters in $\{\bm{W}_{p,q}\}$ can get very large, we restrict each $\bm{W}_{p,q}$ to be a diagonal matrix, which can equivalently be represented by the vector $\bm{w}_{p,q}$. It is similar to a depth-wise convolution, but the weights are not shared between different positions in the feature map. Equation \ref{eq:modifiedconv} can now be rewritten as
\begin{equation}
\bm{y}_p = \sum_{q\in\mathcal{R}} \bm{w}_{p,q} \otimes \bm{x}_{p+q}.\label{eq:modifiedconv2}
\end{equation}
where $\otimes$ represents a component-wise (Schur) multiplication. The filter parameters $\bm{w}_{p,q}$ are derived from $\bm{X}$ through a convolution operation as 
\begin{equation}
\bm{w}_{p,q} = \sum_{t\in\mathcal{R'}}\Theta_{t}^{q}\bm{x}_{p+t}
\label{eq:fulltheta}
\end{equation}
where $\mathcal{R'}$ (similar to $\mathcal{R}$) is the receptive field for the convolutional filters, and represents the {\em context field}, i.e., the span from which contexts are captured to compute the weight $\bm{w}_{p,q}$, and $\Theta_t^{q}$ are the convolutional operation's parameters. Each $\Theta_t^q$ is a $C_{in}\times C_{in}$ matrix.  The complete set of parameters of $g$ are given by $\Theta = \{\Theta^{q}_t,~q \in \mathcal{R},~t \in \mathcal{R'} \}$, with the total number of parameters equal to $C_{in}^2\times |\mathcal{R}| \times|\mathcal{R'}|$ where $|\mathcal{R}|$ is the number of elements in $\mathcal{R}$. 

The shared weights $\Theta$ capture the higher-level patterns required to characterize spatio-temporal context. We define $\mathcal{R}'$ as the $\textbf{context field}$ where context information is captured and define $\mathcal{R}$ as the $\textbf{kernel size}$ where features are extracted from $|\mathcal{R}|$ positions.

\subsection{ConvNet-based second-order operation}
In the ConvNet-based second-order block, we use a small convolutional neural network comprising multiple layers of convolutions followed by activations to compute $\bm{w}_{p,q}$, i.e. $\bm{w}_{p,q}=\text{ConvNet}_q\left(\{\bm{x}_{p+t},t\in\mathcal{R}'\};\Theta^q\right)$. We can generate all $\bm{w}_{p,q}$ with one ConvNet. Representing $\bm{w}^p = \{\bm{w}_{p,q},~ q\in\mathcal{R}\}$, we can write
\begin{equation}
\bm{w}^{p} = \text{ConvNet}\left(\{\bm{x}_{p+t},t\in\mathcal{R}'\};\Theta\right),\label{eq5}
\end{equation}
where $\Theta$ are the parameters of the ConvNet. 

Though ConvNets consist of multiple layers of convolutions, the number of parameters in Equation \ref{eq5} can be smaller than that required by the convolutional-based model of Equation \ref{eq:fulltheta}. Typically we need a non-local context field, so $\mathcal{R}'$ is relatively bigger than the common convolutional kernel size. For instance, we want a $5\times5\times5$ context field. A ConvNet with three layers, each computed by a $1\times3\times3$, $3\times3\times3$ and $3\times1\times1$ filter using only 39 parameters, whereas a single convolution would require a $5\times5\times5$ filter with 125 parameters to provide the same context field.  Furthermore, a ConvNet with non-linear activations can have better representation abilities, and capture more completed higher-order structures.

\textbf{H-blocks.}  Following \cite{qiu2017learning}, we use 3 layers of Pseudo-3D (P3D) convolutions to implement the ConvNet $\text{ConvNet}(\cdot; \Theta)$ in Equation \ref{eq5} for obtaining a sufficiently large context field. Table \ref{tb0} shows the kernel size of three P3D convolutions as the factorization of different context fields.
\begin{table}[h]
\centering
\begin{tabular}{@{}cccc@{}}
\toprule
 context field & layer 1 & layer 2 & layer 3 \\ \midrule
$3\times3\times3$& $1\times3\times3$& $3\times1\times1$& $1\times1\times1$    \\
$3\times5\times5$& $1\times3\times3$& $3\times3\times3$& $1\times1\times1$    \\
$5\times5\times5$& $1\times3\times3$& $3\times3\times3$& $3\times1\times1$    \\
$5\times7\times7$& $1\times3\times3$& $3\times3\times3$& $3\times3\times3$    \\ 
$7\times7\times7$& $3\times3\times3$& $3\times3\times3$& $3\times3\times3$    \\ \bottomrule
\end{tabular}
\caption{Factorization of different context fields. For example, we stack three convolutions with kernel size $3\times3\times3$ to get a $7\times7\times7$ context field.}
\label{tb0}
\end{table}

Suppose the number of the H-block's input channels is $C$ and the kernel size of the H-block is $|\mathcal{R}|$, the number of input channels and output channels for the three P3D convolutions are ($C$, $C$), ($C$, $C//|\mathcal{R}|\times |\mathcal{R}|$) and ($C//|\mathcal{R}|\times |\mathcal{R}|$,$C\times|\mathcal{R}|$) respectively ($//$ is integer division, for example $19//9=2$).
After each convolution layer, we use the scaled exponential linear unit (SELU) \cite{klambauer2017self} as the activation. The last convolution is always a group convolution \cite{xie2017aggregated} with group size $|\mathcal{R}|$ to reduce parameters. And we use \emph{softmax} as the last convolution's activation as a normalization factor.

\section{Experiments}
We perform comprehensive studies on the challenging Something-Something V1 dataset \cite{sth-sth-v2}, and also report results on the Charades \cite{charades}, Kinetics-400 \cite{i3d} and Something-Something V2 dataset to show the generality of our models.

\subsection{Implementation Details}
To draw fair comparison with the results in \cite{Wang_nonlocalCVPR2018, wang_gcn} on the same datasets, our backbone model is based on the ResNet-50 Inflated 3D architecture (Table \ref{backbone}) and is the same as that in \cite{wang_gcn} . Note there are small differences between our backbone model with the Inflated 3D backbone in \cite{Wang_nonlocalCVPR2018} where
the output of the last convolutional layer is a $\sfrac{T}{2}\times 14\times14$ feature map ($T$ is the number of input frames).
   \begin{table}[ht]
   \centering
        
      \begin{tabular*}{8.5cm}{c|c|c}
         \hline
         \multicolumn{2}{c|}{layer} & output size\\
         \hline
         \begin{minipage}{1cm}\vspace{2mm} \centering conv$_1$ \vspace{1mm} \end{minipage} &$5\times7\times7$, 64, stride 1,2,2& $T\times 112\times112$\\
         \hline
         \begin{minipage}{1cm}\vspace{2mm} \centering pool$_1$ \vspace{1mm} \end{minipage} &$1\times3\times3$, max, stride 1,2,2& $T\times 56\times56$\\
         \hline
         \begin{minipage}{1cm}\vspace{7mm} \centering res$_2$ \vspace{7mm} \end{minipage} &$\begin{bmatrix}3\times1\times1,64\\1\times3\times3,64\\1\times1\times1,256\end{bmatrix}\times3$& $T\times 56\times56$\\
         \hline
         \begin{minipage}{1cm}\vspace{2mm} \centering pool$_2$ \vspace{1mm} \end{minipage} &$3\times1\times1$, max, stride 2,1,1& $\frac{T}{2}\times 56\times56$\\
         \hline
         \begin{minipage}{1cm}\vspace{7mm} \centering res$_3$ \vspace{7mm} \end{minipage} &$\begin{bmatrix}3\times1\times1,128\\1\times3\times3,128\\1\times1\times1,512\end{bmatrix}\times4$& $\frac{T}{2}\times 28\times28$\\
         \hline
         \begin{minipage}{1cm}\vspace{7mm} \centering res$_4$ \vspace{7mm} \end{minipage} &$\begin{bmatrix}3\times1\times1,256\\1\times3\times3,256\\1\times1\times1,1024\end{bmatrix}\times6$& $\frac{T}{2}\times 14\times14$\\
         \hline
         \begin{minipage}{1cm}\vspace{7mm} \centering res$_5$ \vspace{7mm} \end{minipage} &$\begin{bmatrix}3\times1\times1,512\\1\times3\times3,512\\1\times1\times1,2048\end{bmatrix}\times3$& $\frac{T}{2}\times 14\times14$\\
         \hline
         \multicolumn{2}{c|}{global average pool and fc}& \begin{minipage}{1cm}\vspace{2mm} 1$\times$1$\times$1\vspace{1mm}\end{minipage}\\
         \hline
      \end{tabular*}\\
      \caption{Our backbone ResNet-50 I3D model. We use T$\times$H$\times$W to represent the dimensions of kernels and output feature maps. $T=\{8,32\}$, and the corresponding input size is 8$\times$224$\times$224 and 32$\times$224$\times$224.}
      \label{backbone}

   \end{table}

\textbf{Training.} Unless specified, all the models are trained from scratch. Following \cite{wang_gcn}, we first resize the input frames to the $256\times 320$ dimension and then randomly crop $224\times224$ pixels for training. We first train our model with 8-frame input clips randomly sampled in 12 frames per second (FPS) on a 4-GPU machine with a batch size of 64 for 30 epochs, starting with a learning rate of 0.01 and reducing it by a factor of 10 at $15^{\text{th}}$ epoch. Then we fine-tune the model with 32-frame input randomly sampled in 6FPS on an 8-GPU machine with a batch size of 32 for 45 epochs, starting with a learning rate of 0.01 and reducing by a factor of 10 at every 15 epoch.

We use mini-batch stochastic gradient descent with a momentum of 0.9 and a weight decay of 1e-4 for optimization. We use cross entropy loss function for Something-Something V1, V2 and Kinetics-400 datasets, and binary sigmoid loss for Charades datasets (multi-class and multi-label).

\textbf{Inference.}   At the inference stage, we resize the input frames to the $256\times 320$ dimension, randomly sample 40 clips of 32-frame inputs in 6FPS, randomly crop $224\times224$ pixels for testing. The final predictions are based on the the averaged softmax scores of 40 all clips.
   
\subsection{Experiments on Something-Something V1}

\begin{table*}[h] 
\begin{minipage}{0.35\textwidth} 
\centering 
\begin{tabular}{@{}ccc@{}}
\toprule
Model & Top-1 & Top-5 \\ \midrule
I3D  ResNet-50 & 41.6  & 72.2  \\
res2  & 43.6  & 74.3  \\
res3  & 43.7 & 74.6  \\
res4  & 43.4  & 74.2  \\
res4  & 42.1  & 73.5  \\ \bottomrule
\end{tabular}
\end{minipage}
\begin{minipage}{0.35\textwidth} 
\centering 
\begin{tabular}{@{}ccc@{}}
\toprule
Model & Top-1 & Top-5 \\ \midrule
I3D ResNet-50   & 41.6  & 72.2  \\
res3-1  & 43.6  & 74.4  \\
res3-2  & 43.7  & 74.6  \\
res3-3  & 43.3  & 74.2  \\
res3-4  & 42.9  & 74.0  \\ \bottomrule
\end{tabular}
\end{minipage}
\begin{minipage}{0.35\textwidth} 
\centering 
\begin{tabular}{@{}ccc@{}}
\toprule
Model & Top-1 & Top-5 \\ \midrule
I3D ResNet-50  & 41.6  & 72.2  \\
1-block  & 43.7  & 74.2  \\
3-block  & 46.2  & 76.1  \\
5-block  & 48.6 & 78.1  \\ \bottomrule
\end{tabular}
\end{minipage}
\\

\begin{minipage}{0.3\textwidth} 
(a) Stages: 1 H-block is added into different stages (with same context field of 5x5x5 and same learnable kernel of 3x3x3).
\end{minipage}
\qquad\qquad
\begin{minipage}{0.3\textwidth}
(b) Position within stages: 1 H-block is added into different positions of in the same stage (with same context field of 5x5x5 and same learnable kernel of 3x3x3).
\end{minipage}
\qquad\qquad
\begin{minipage}{0.3\textwidth} 
(c) Number of H-blocks added: 1 block is added at Res3-2. 3 blocks are added at Res3-2, Res4-2 and Res4-4. 5 blocks are added to every other residual block of Res3 and Res4.
\end{minipage}
\quad\\

\begin{minipage}{0.35\textwidth} 
\centering 
\begin{tabular}{@{}ccc@{}}
\toprule
Model & Top-1 & Top-5 \\ \midrule
I3D ResNet-50   & 41.6  & 72.2  \\
$3\times1\times1$  & 48.0  & 77.1  \\
$1\times3\times3$  & 48.1  & 77.3  \\
$3\times3\times3$  & 48.6  & 78.1  \\
$3\times5\times5$  & 48.3  & 77.6  \\ \bottomrule
\end{tabular}
\end{minipage}
\begin{minipage}{0.35\textwidth} 
\centering 
\begin{tabular}{@{}ccc@{}}
\toprule
Model & Top-1 & Top-5 \\ \midrule
I3D ResNet-50  & 41.6  & 72.2  \\
$3\times5\times5$  & 48.2  & 77.3  \\
$5\times5\times5$  & 48.6  & 78.1  \\
$5\times7\times7$  & 48.5  & 77.6  \\
$7\times7\times7$  & 48.2  & 77.6  \\ \bottomrule
\end{tabular}
\end{minipage}
\begin{minipage}{0.35\textwidth} 
\centering 
\begin{tabular}{@{}ccc@{}}
\toprule
Model & Top-1 & Top-5 \\ \midrule
I3D ResNet-50  & 41.6  & 72.2  \\
softmax  & 48.6  & 78.1  \\
relu  & 48.3  & 77.7 \\
tanh  & 48.4  & 77.9  \\ \bottomrule
\end{tabular}
\end{minipage}
\quad\\

\begin{minipage}{0.3\textwidth} 
(d) Learnable kernels: 5 H-blocks are added with different learnable kernels (with same context field 5x5x5). 
\end{minipage}
\qquad\qquad
\begin{minipage}{0.3\textwidth} 
(e) Context Field: 5 H-blocks are added with different context field (with same learnable kernels 3x3x3). 
\end{minipage}
\qquad\qquad
\begin{minipage}{0.3\textwidth} 
(f) Activations: 5 H-blocks are added with different activations (with same context field of 5x5x5 and same learnable kernel of 3x3x3).
\end{minipage}

\caption{Ablations on Something-Something V1 action classification. We show top-1 and top-5 classification accuracy} 
\label{table:ab} 
\end{table*}

   Something-Something V1 dataset has 86K training videos, around 12K validation videos and 11K testing videos. The number of classes in this dataset is 174. 
   
   Table \ref{table:ab} shows the ablation results on the validation dataset, analyzed as follows:
   
   \textbf{Higher-order at different stages.} We study the network performance when the H-blocks are added to different stages on the network. We add one single H-block after the first bottleneck on 1) res2, 2) res3, 3) res4 and 4) res5 in Table \ref{backbone}. As shown in Table \ref{table:ab}, the improvement of adding one H-block on res3 is the most prominent. The improvement decreases when adding the H-block to deeper stage of the network. One possible explanation is that spatiotemporal correlation weakens as the network getting deeper, since high level features are more linearly separable so $\emph{higher-order}$ information is less important. One possible reason that $\emph{higher-order}$ on res2 cannot get the maximum improvement is that the output size of res2 is 8 times larger than the output size of res3, thus the context field is much smaller compared with the entire feature map. An evidence can be found in the following study.
   
   \textbf{Higher-order at different positions of the same stage.} We further discuss the performance of adding a single H-block to different positions of the same stage. We add one single H-block after 1) first, 2) second, 3) third and 4) fourth bottleneck within res3. From Table \ref{table:ab}, We find that adding one H-block after the first and second bottleneck (res3-1 and res3-2) leads to a better accuracy than adding the H-block in res3-3 and res3-4. This again proves that spatiotemporal contexts weakens as the network going deeper, and our single H-block can capture more meaningful spatiotemporal contexts and lose less information than deep stack of convolution layers.

	\begin{table*}[ht]
	\centering
		\begin{tabular}{l|c|c|c|c|c}
		\hline
		Method & Pre-train dataset &  Input size & Backbone & Modality & Top1 Acc.(\%) \\
		\hline
		MultiScale TRN \cite{trn_eccv2018} & Imagenet & - & Inception & RGB & 33.6 \\
			ECO \cite{ECO_eccv18} & - & multi-input ensemble & Inception+3D ResNet 18 & RGB+Flow & 43.9 \\
		
		I3D \cite{i3d} & ImageNet,Kinetics & $16\times224\times224$ & ResNet 50 & RGB & 41.6 \\
		NL I3D \cite{Wang_nonlocalCVPR2018} & ImageNet,Kinetics & $32\times224\times224$ & ResNet 50 & RGB & 44.6 \\
		NL I3D + GCN \cite{wang_gcn} & ImageNet,Kinetics & $32\times224\times224$ & ResNet 50 & RGB & 46.1 \\
\hline
		HO I3D [ours] & None & $32\times224\times224$ & ResNet 50 & RGB & 45.7 \\
		HO I3D [ours] & ImageNet,Kinetics & $32\times224\times224$ & ResNet 50 & RGB & 48.2 \\
		\hline
		\end{tabular}
	\caption{Test results on Something-Something V1 dataset. NL is short for non-local.}
	\label{sth-v1-big-table}
	\end{table*}
   
\textbf{Going deeper with H-blocks.} Table \ref{table:ab} shows the results of adding more higher-order blocks. We add 1 block, 3 blocks (2 to Res3 and 1 to Res4) and 5 blocks (3 to Res3 and 2 to Res4) in ResNet-50. More H-blocks in general lead to better results. We argue that multiple higher-order blocks can capture comprehensive contextual information. Messages in each location can be learned with its own context, which is hard to do via shared weights.

\textbf{H-blocks within different kernel sizes.} We study how the kernel size would influence the improvement by adding 5 blocks of H-blocks with different kernel sizes and same context field ($5\times5\times5$). As shown in Table \ref{table:ab}, H-blocks with a kernel size of $3\times3\times3$ are the best. Smaller and larger will lower the classification accuracy. The reduced performance for the $3\times5\times5$ may come from the difficulty in optimizing the network due to the larger spatial size.
\begin{figure*}[htp]
\includegraphics*[width=1\textwidth]{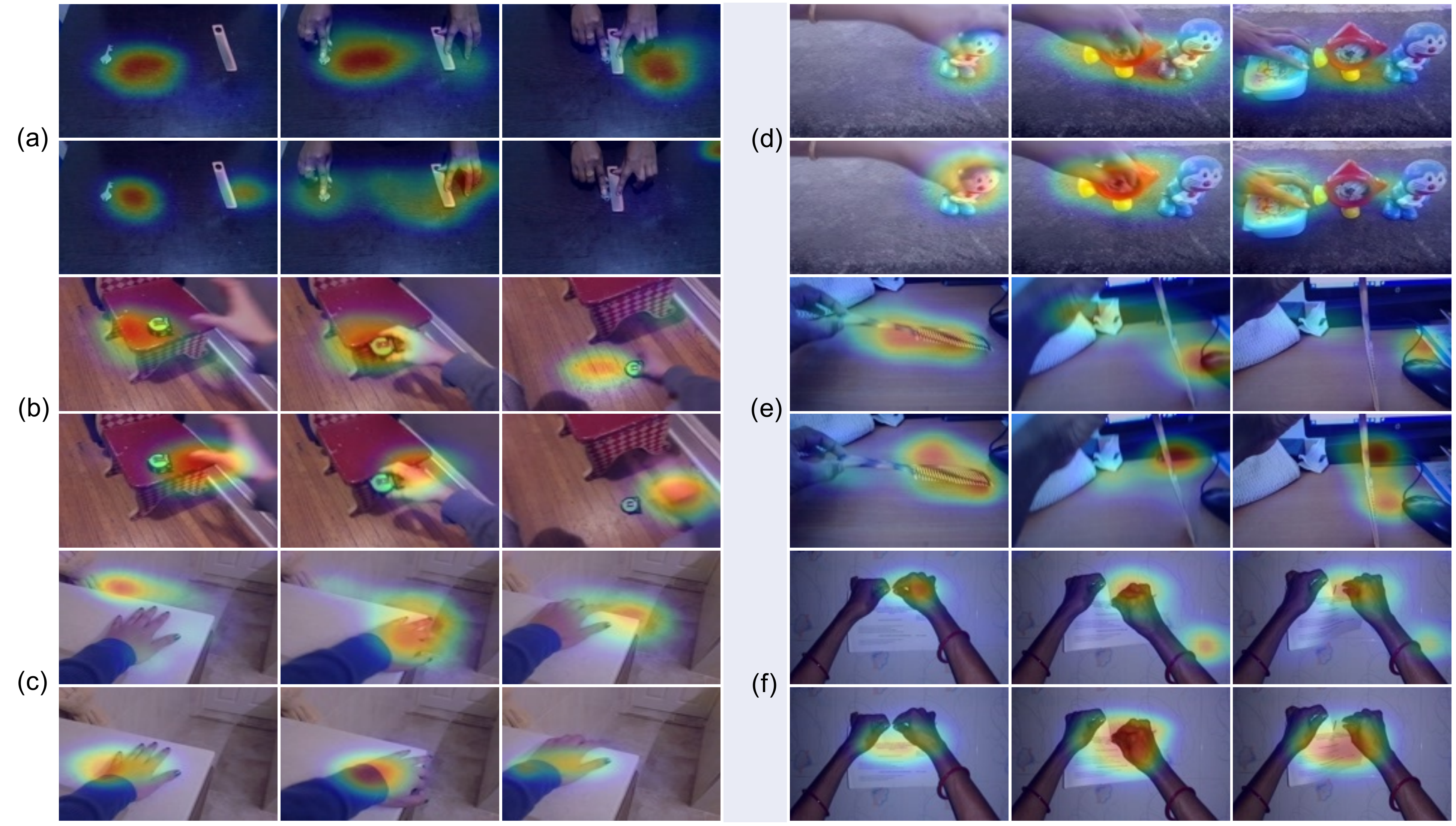}
\caption{Visualizing learned feature map. The upper row of each sample is the feature map of I3D, the bottom row is the feature map of the higher-order network. Videos are from Something-Something V1 dataset, with labels of: (a) Moving something and something closer to each other; (b) Moving something down; (c) Touching (without moving) part of something; (d) Putting something, something and something on the table; (e) Lifting up one end of something without letting it drop down; (f) Tearing something just a little bit.}
\label{vis}
\end{figure*}

\textbf{H-blocks with different context fields.} We study how the size of context fields influence the improvement by adding 5 blocks of H-blocks with different context fields and same kernel size ($3\times3\time3$). In Table \ref{tb0}, we show other possible context fields and their factorization using three convolutions. As shown in Table \ref{table:ab}, The improvement of the H-block with context fields of $5\times5\times5$ and  $5\times7\times7$ is similar, and a smaller context field of  $3\times5\times5$ as well as a larger context field of $7\times7\times7$ is slightly smaller. One possible
explanation is that the smaller context field has a small context and it is insufficient to provide precise contextual information. And for larger context field, the context is redundant and more difficult to capture.

\textbf{H-blocks with different activation functions.} Instead of using \emph{softmax}, we also use ReLU and \emph{tanh} as the last activations. As shown in Table \ref{table:ab}, different activation functions versions perform similarly, illustrating that activation function of this module is not
the key to the improvement in our applications; instead, it is more likely that the higher-order behavior is essential, and it is insensitive to the activation functions.

\textbf{Comparison to the state of the art.} We compare the performance with the state-of-the-art approaches on the test set of the Something-Something V1 dataset. The results are summarized in Table \ref{sth-v1-big-table} (HO is short for higher-order). We use a $5\times5\times5$ context field and a $3\times3\times3$ kernel size with the  \emph{softmax} activation function. We get a top-1 accuracy of 45.7\% without pre-training with other image or video datasets. When pre-trained with ImageNet and Kinetics, our model gets a top-1 accuracy of 48.2\%, which is the highest single model result on the leaderboard, surpassing all the existing RGB based methods by a good margin. Note that there are a few related works unlisted such as \cite{crasto2019mars}. They used a deeper backbone such as ResNet101, more input modalities and more input frames, thus are not comparable with our results.

The higher-order operation is a light and efficient module that significantly improves the network performance with limited extra computational cost. If given the 32 frames as the input, I3D model \cite{i3d}, non-local I3D model \cite{Wang_nonlocalCVPR2018}, and our HO I3D model using the ResNet50 backbone (Table \ref{backbone}) take 326G, 401G, and 368G FLOPs respectively.

Figure \ref{vis} visualizes several examples of the feature maps learned by our H-blocks block as well as the I3D ResNet-50 backbone. All the feature maps are from the output of the res5 stage in Table \ref{backbone}, resized back to the size of the original videos. In Figure \ref{vis}(a) \emph{moving something and something closer to each other}, our model is focusing simultaneously on two objects and the hands, showing that our model can not only capture appearance information but also capture motion information. In Figure \ref{vis}(d) \emph{putting something, something and something on the table}, we can see evident differences between I3D and H-blocks in the third frame, in which I3D is looking at the red clock, while H-block is focusing on the moving part - hand. From Figure \ref{vis}, we can conclude that our higher-order network can learn to find important relation clues instead of focusing on appearance information compared with I3D backbones. 

We also investigate our models on Something-Something V2 dataset. The V2 dataset has 22K videos, which is more than twice as many videos as V1. There are 169K training videos, around 25K validation videos and 27K testing videos in the V2 dataset. The number of classes in this dataset is 174, which is the same as the V1 version. Table \ref{sth-v2-table} shows the comparisons with the previous results on this dataset. When adding five higher-order blocks to res3 and res4 stages, our higher-order ResNet 50 achieves 62.6\% Top 1 accuracy. 

	\begin{table}[ht]
		\begin{tabular}{p{3.5cm}|p{1.5cm}|p{2cm}}
		\hline
			model & backbone  &  Top1 Acc.(\%) \\
			\hline
         Multi-Scale TRN \cite{trn_eccv2018}   & Inception& 48.8 \\
         2-Stream TRN \cite{trn_eccv2018}   & Inception&  56.5 \\
			\hline
			HO I3D [ours]  & ResNet 50 & 62.6 \\
		\hline
		\end{tabular}
		\caption{Validation results on Something-Something V2 Dataset.}
		\label{sth-v2-table}
	\end{table}

Kinetics-400 \cite{i3d} contains approximately 246k training videos and 20k validation videos. It is a classification task involving 400 human action categories. We train all models on the training set and test on the validation set.

Table \ref{kinetics-big-table} shows the comparisons with the state-of-art on this dataset. We use the best settings from section 4.2, which is 5 H-blocks with $5\times5\times5$ context field, $3\times3\times3$ kernel size and \emph{softmax} activation. Our model achieve a top-1 accuracy of 77.8 and top-5 accuracy of 93.3. Compared with methods that use RGB and Flow, our method can learn motion information end-to-end. Our model is also better than those using RGB only for training.

\subsection{Experiments on Kinetics-400}
   \begin{table}[htp]
   \centering

      \begin{tabular}{l|c|c|c}
      \hline
      Method & Backbone & Top-1&Top-5  \\
      \hline
      ARTNet \cite{wang2017appearance} & ResNet 18 & 69.2 & 88.3 \\
      I3D \cite{i3d} & BN-Inception &  71.1 & 89.3 \\ 
      2-stream I3D \cite{i3d} & BN-Inception &  74.2 & 91.3 \\ 
      2-stream R(2+1)D \cite{r2plus1d} & ResNet 50  & 73.9 &90.9 \\
      NL I3D \cite{Wang_nonlocalCVPR2018} & ResNet 50  & 76.5 & 92.6 \\
      NL I3D \cite{wang_gcn} & ResNet 101 & 77.7 & 93.3 \\
      SlowFast \cite{slowfast} & ResNet 50  & 77.0 & 92.6 \\
      NL SlowFast\cite{slowfast} & ResNet 50 & 77.7 & 93.1 \\
      
\hline
      HO I3D [ours] & ResNet 50 &  77.8 & 93.3 \\
      \hline
      \end{tabular}
      \caption{Validation results on Kinetics-400 dataset }
   \label{kinetics-big-table}
   \end{table}

\subsection{Experiments on Charades dataset}
In this subsection we study the performance of higher-order neural networks on the Charades dataset. 
The Charades dataset is a dataset of daily indoors activities, which consists of 8K training videos and 1.8K validation videos. The average video duration is 30 seconds. There are 157 action classes in this dataset and multiple actions can happen at the same time. We report our results in Table \ref{charades-table}. The baseline I3D ResNet 50 approach achieves 31.8\% mAP. The best result NL I3D + GCN \cite{wang_gcn} in Table \ref{charades-table} is a combination of two models. By adding 2 H-blocks to res3 and and 3 to res4 stages in the I3D Res50 backbone, our method archives 5.1\% improvements (36.9\% mAP) in mAP. And we archive another 0.2\% gain (37.1\% mAP) by continuously adding 2 H-blocks to res2 stage. The improvement indicates the effectiveness of H-blocks.

\begin{table}[ht]
	\begin{tabular}{p{4cm}|p{2cm}|p{0.8cm}}
	\hline
		model & backbone  & mAP \\
		\hline
		    Two-Stream \cite{twostream_cvpr2017} & VGG 16 &  18.6 \\
		    MultiScale TRN \cite{trn_eccv2018} & Inception  & 25.2 \\
			I3D \cite{i3d}  & ResNet 50 & 31.8 \\
			I3D \cite{i3d}  & Inception & 32.9 \\
			I3D \cite{Wang_nonlocalCVPR2018} & ResNet 101 & 35.5 \\
			NL I3D \cite{Wang_nonlocalCVPR2018}  & ResNet 50  & 33.5 \\
			GCN \cite{wang_gcn} & ResNet 50  & 36.2 \\
			NL I3D + GCN \cite{wang_gcn} & ResNet 50  &37.5\\
			\hline
			HO I3D [ours]  & ResNet 50 & 37.1 \\
	\hline
	\end{tabular}
	\caption{Validation results on the Charades dataset \cite{charades}. NL indicates Non-Local.}
	\label{charades-table}
\end{table}

\section{Conclusion}
In this paper, we have introduced higher-order networks to the task of action recognition. Higher-order networks are constructed by a general building block, termed as  H-block, which aims to model position-varying contextual information. As demonstrated on the Something-Something (V1 and V2), Kinetics-400 and Charades datasets, the proposed higher-order networks are able to achieve state-of-the-art results, even using only RGB mobility inputs without fine-tuning with other image or video datasets. The good performance may be ascribed to the fact that higher-order networks are a natural for context modeling. 

The actual model itself is not restricted to visual tasks, but may be applied in any task where a context governs the interpretation of an input feature, such as cross-modal or multi-modal operations. In future work, we plan to investigate the benefits of our higher-order model and its extensions, in a variety of other visual, text and cross-modal tasks.

{\small
\bibliographystyle{ieee_fullname}
\bibliography{egbib}
}

\end{document}